\documentclass[12pt]{article}
\usepackage[usenames]{color}
\usepackage{latexsym}
\usepackage{amsmath,amsthm,amssymb}
\usepackage{hyperref} 

 \advance\textwidth  by 5.0cm
 \advance\textheight by 6cm

\newtheorem{proposition}{Proposition}

\theoremstyle{definition}
\newtheorem{definition}{Definition}

\theoremstyle{remark}
\newtheorem{remark}{Remark}

\theoremstyle{property}

\title{Genetic algorithms in Forth}

\author{Khashin~S.~I., Vaganov~S.~E. \\
        Ivanovo State University, Russia \\
        \texttt{khash2@gmail.com} }
\begin{document}

\maketitle

\tableofcontents

\begin{abstract}
A method for automatically finding a program (bytecode) realizing
the given algorithm is developed. The algorithm is specified as a
set of tests (input\_data) $ \rightarrow $ (output\_data). Genetic
methods made it possible to find the implementation of relatively
complex algorithms: sorting, decimal digits, GCD, LCM, factorial,
prime divisors, binomial coefficients, and others. The algorithms
are implemented on a highly simplified version of Forth language.

\end{abstract}

\section{Introduction}

We want to automatically receive a program in some programming
language that implements an algorithm, defined as a set of tests. To
construct such an algorithm, we will use methods that in some sense
have analogies in biology, in genetics. This approach is called
"genetic programming" \cite{genProg,arh17,wikiLGP}.

These ideas first appeared in the 60s of the last century
\cite{wikiLGP}, but remains quite popular until today. At least at
the site scholar.google.com search for the words "genetic algorithm"
for 2017/18 year gives more than $46\,000$ links.

For example, the site \cite{genProg} is devoted to genetic
programming, it describes the results obtained.

There exist a special program for finding algorithms using the
genetic methods \cite{SlashA}, but examples of algorithms found with
its help are known only to the simplest ones (function $x
\rightarrow x^2 + x + 1$).

The work \cite{arh17} describes the results achieved on this way.
Authors managed to find programs that implement the following
functions:
\begin{itemize}
\item nguen1: $x \rightarrow x^3 + x^2 + x$,
\item nguen2: $x \rightarrow x^4 + x^3 + x^2 + x$,
\item nguen3: $x \rightarrow x^5 + x^4 + x^3 + x^2 + x$,
\item nguen4: $x \rightarrow x^6 + x^5 + x^4 + x^3 + x^2 + x$,
\item keijzer4: $x \rightarrow x^3e^{-x}\cos(x)\sin(x)(\sin^2(x)\cos(x)-1)$,
\end{itemize}
and several other functions of this kind.

In the book \cite{Sheppard} methods of genetic programming in Python
are considered. Here, more than a dozen different problems are
solved by genetic methods. However, the main difficulty in them is
not the genetic algorithms themselves, but the suitable
reformulation of the problem. After that, the problem can be solved
by the genetic method, or directly.

The widely distributed machine learning package "TensorFlow"\,
\cite{TensorFlow} also contains a tools for genetic programming. You
can read about the results achieved with it, for example, in
\cite{arh17a}.

In fact, in almost cases, the considered "genetic" methods are
reduced to the search for a minimum of some loss function from some,
possibly, quite a considerable number of parameters (several dozen).

In any case, this approach takes us far enough from the initial idea
of a genetic approach. In this paper we will try to stay within the
framework of the originally formulated task.

\section{Problem definition}

We want to find a program in some programming language implementing
given algorithm. The algorithm will be defined as a set of tests:
(input\_data) $\rightarrow$ (output\_data).

The unit of the test is the pair $t=(x,y)$, where $(x \in X, y \in
Y)$, and $X,Y$ are the corresponding sets of input and output
objects. In the future, we will consider as input and output objects
arbitrary finite sequences of integers, more precisely, of type int,
32-bit signed numbers.

\subsection{Test files}

A test will be called an arbitrary finite set of test items
$T=(t_0,\dots, t_{n-1})$.

Sometimes, when there are no confusion, we will call a single test
element as a test.

In practice, the test is represented by a text file of the form :
\begin{verbatim}
 #T SHIFTR_0 x,y ->  x*2^y  comment
 <in> 9   3 </in><out>  1536 </out>
 <in> 4   3 </in><out>    48 </out>
 <in> 2   7 </in><out>    28 </out>
 <in> 3   3 </in><out>    24 </out>
 <in> 6   5 </in><out>   320 </out>
 <in> 2   5 </in><out>    20 </out>
 <in> 7   5 </in><out>   640 </out>
 ...
\end{verbatim}

The first two characters {\tt \#T} are a signature of the test file,
the next word {\tt SHITFR\_0} is the recommended name of the program
you are looking for, the rest of the line is a comment.

In each of the following lines, a set of input and output integers
parameters of the desired algorithm is defined.

We want to find a program that from each set of input data generates
the required output set.

\begin{definition} Pair $(m,n)$, where $m$ is a number of input integers and
$n$ is a number of output integers will be called the signature of
the test element.
\end{definition}

\begin{remark} In general, it is allowed to have test elements with
various signatures in one test. But this possibility was not used in
current work.
\end{remark}

The number of test items in each test is from several units to
hundreds.

A large set of test files is prepared for testing the developed
methods, they are available on the site \cite{gene_exe}. Not for all
of them was found implementing programs. Tests are collected by
sections:

\begin{itemize}
\item Polynomialы;
\item Max/min and sorting;
\item GCD, LCM, factorials;
\item Number theory;
\item Decimal and binary digits;
\item Power;
\item $sqrt, log, \dots $;
\item Interval testing;
\item Rest;
\end{itemize}

Some tests have multiple version: complete and simplified. For
example, for GCD ({\tt GCD.tst}) there is a test {\tt GCD\_0.tst}
containing only positive input parameters. Such simplified tests are
required in the case when the complete test can not be implemented
immediately.

\section{Programming language}

Programs from our point of view are of two types: linear and
structural. The advantages and disadvantages of each approach are
quite obvious and there is no fundamental difference between them
from our point of view. We have chosen a linear approach in our
work, at least for the initial stage.

Usually, this is done by selecting an assembly-type instruction set
(registers). But we decided that the stack approach, actually
implemented in the language of Forth \cite{thinking-forth,
wiki_Forth}, would be more effective.

\subsection{Forth}

To find the program with the genetic method, the most stripped-down
version of the Forth language was chosen. Only those instructions
are left, which are impossible to do without.

There is no interaction with the user,  such as output to the
screen, even variables are missing. The control structures are
represented only by unconditional and conditional jump-instruction.

Language Forth is very compact, new words (functions) are introduced
very easily. In Forth, the newly defined functions have exactly the
same syntax as the built-in language elements. This is convenient
for a genetic approach. Thanks to this, you do not need to go
through programs with complex structure in the form of a tree, only
the simplest ones.

An example of a program that calculates the sum of the squares of
two numbers and the factorial:

\begin{verbatim}
 : SUMSQ2 DUP * SWAP DUP * + ;
 : FACTORIAL CONST 1 OVER -- -ROT * OVER IF -6 SWAP DROP ;
\end{verbatim}

In this paper, we will call the language a Forth for short, although
we must remember that this is a strongly truncated version of the
language.

\subsection{Structure of the program}

All the programs and functions  will work with the data stack. Only
4-byte integers are stored on the stack. The functions in the Forth
have no arguments. The initial data they take from the stack and
there they leave the results of their work.

We call the signature of a function a pair of nonnegative numbers
$(p,q)$, which show the number of input and output arguments of the
function in the stack.

The number of arguments can vary, depending on the state of the
stack. Such a signature will be called "floating". For now, we will
only consider functions with a fixed, static signature. At each
point of such a program, the current stack depth is statically
determined.

Later, functions will be called "words", or "genes". At each moment
of time, the system has a certain set of built-in words (functions)
and the current set of new genes, that is, new words built in the
learning process ("evolution"). One step of evolution is the
construction of one or more new words (functions) that solve one
problem (test) in whole or in part.

\subsection{Bytecode}

The program in the Forth is a byte sequence, each byte is a separate
command. There are two types of commands: built-in and implemented
on the Forth itself. Thus, the total number of commands can not be
more than $255$.

In our version of Forth there are  $33$ built-in commands:

$0$. A sign of the end of a function or program. There should be no
other zero bytes in the function (almost). Thus, the function can be
considered as a NULL-terminated string.

$1$: Unconditional jump to the specified address. The command
consists of 2 bytes: the command code (1) and the next byte
indicating the relative address of the jump ($-128 \dots 127$). The
functions considered in the system will be short, so the restriction
on the amount of displacement is not essential. For example, in the
code
\begin{verbatim}
  OVER + GOTO 2 DUP - SWAP
\end{verbatim}
the command "GOTO 2" means jump to the command "-", and in the code
\begin{verbatim}
 ROT OVER + GOTO -3 DUP - SWAP
\end{verbatim}
command "GOTO -3" means jump to the command "OVER".

The transition with $0$ offset does not make sense, so the second
byte of the command is always non-zero.

$2$: Conditional branch to the specified address. The command
consists of $2$ bytes: the command code ($2$) and the next byte
indicating the relative jump address ($-128 \dots 127$). The second
byte of the command is again always nonzero.

$3$: Numeric literal: put a number from interval $-128\dots 127$ on
the stack. The command consists of 2 bytes: code (3) and byte-value.
If we do not want to use the command "CONST 0" containing zero byte,
we can add the word "ZERO" to the dictionary, in this case the
second byte of the command is always nonzero and the program will
not contain zeroes, except for the final zero, treated as
"NULL-terminated string".

The three commands described are two bytes, all others consist of
one byte.

Stack manipulation commands.
\begin{verbatim}
  Command     Stack before     Stack after
  DUP         a                a a
  DROP        a
  SWAP        a b              b a
  OVER        a b              a b a
  ROT         a b c            b c a
  -ROT        a b c            c a b
\end{verbatim}

In addition, to manipulate the stack, there are two commands "PICK"
and "ROLL":

"PICK": get the $n$-th number, where $n$ is the top of the stack,
for example: "0 PICK" is equivalent to "DUP", "1 PICK" is equivalent
to "OVER".

"ROLL": rotate $n$ numbers, for example: "1 ROLL" is equivalent to
"SWAP" , "2 ROLL" is equivalent to "ROT".

These functions are not static: the required depth of the stack
depends on the value of its vertex. In order to use only commands
with a static signature, the following five static versions are
added:

\begin{verbatim}
  Command     Stack before    Stack after
  2PICK       a b c           a b c a
  3PICK       a b c d         a b c d a
  4PICK       a b c d e       a b c d e a
  3ROLL       a b c d         b c d a
  4ROLL       a b c d e       b c d e a
\end{verbatim}

Arithmetic and bit commands take their arguments on the stack and
leave the result there.

Arithmetic: NEGATE\ \ +\ \  -\ \  *\ \  /\ \  \%\ \ /\% ++ \ \
\verb"--"

Bit commands: AND\ \ OR\ \ XOR\ \ NOT

Logical (comparison):  $>\ \  <\ \ =\ \ 0=\ \ >0\ \  >0$.

In addition to the built-in commands, the system may contain
commands implemented on the Forth itself, for example, finding the
sum of squares of two elements of the stack:
\begin{verbatim}
 : SUMSQ2        DUP * SWAP DUP * + ;
\end{verbatim}

The colon at the beginning of the command is a sign of the beginning
of a new word, after it the name of the word goes, which will be
used in the future in the system. The semicolon at the end is a sign
of the end of the word (function, program).

The word name can be an arbitrary string of characters that does not
contain spaces, tabs to the end of the line. Individual words are
separated by spaces or tabs or line ends.

An important advantage of Forth is that it is not required to
develop a mechanism for "embedding genes"\,, it already exists in
the language. New words (procedures, functions, "genes") are used in
exactly the same way as built-in ones.

Thus, the current list of words (the dictionary) under biological
interpretation will be considered as a "genome".

\section{The first results}

Just try to go through all of programs are not too long.

When examining all programs, we immediately discard programs
containing "forbidden"\, pairs of commands, such as {\tt (DUP DROP)}
or {\tt (SWAP SWAP)}. It should be noted that if the second command
of the forbidden pair has a jump, then such a pair must be resolved.

Even with such a simple approach, it is possible to find the
implementation of some algorithms:

\bigskip

\begin{tabular}{ll}
 Squaring  & {\tt : SQUARE DUP * ; }  \\
 Multiply by $2$ & {\tt  : MUL2 DUP + ;} \\
 Test odd & {\tt : ODD CONST 1 AND ; } \\
 Signum & {\tt : SIGN DUP 0= IF 6 0> DUP IF 2 -- ; } \\
 Sum of suares & {\tt : SUMSQ2 DUP * SWAP DUP * + ; } \\
 Absolute value & {\tt : ABS DUP 0> IF 2 NEG ; } \\
 Descending sort of $2$ numbers & {\tt : SORT2R OVER OVER > IF 2 SWAP ; } \\
 Ascending sort  of $2$ numbers & {\tt : SORT2 DUP 2PICK > IF 2 SWAP ; } \\
 Maximum  of $2$ numbers & {\tt : MAX2 OVER OVER > IF 2 SWAP DROP ; } \\
 Minimum  of $2$ numbers & {\tt : MIN2 DUP 0> IF 2 SWAP IF 2 ROT ; } \\
 Maximum  of $3$ numbers & {\tt : MAX3 OVER 3ROLL > IF 2 SWAP DROP ; } \\
 Minimum  of $3$ numbers & {\tt : MIN3 ROT 2PICK 0> IF -4 DROP DROP ; } \\
\end{tabular}

\subsection{Polynomials}

General polynomials of degrees 1 and 2 are easily:
\begin{verbatim}
 : poly1      * +  ;                 (a,b,x) -> b*x + a
 : poly2      -ROT 2PICK * + * +  ;  (a,b,c,x) -> c*x*x + b*x +a
\end{verbatim}

Specific polynomials
\begin{verbatim}
 : pol1_1    -- DUP + --  ;        x -> 2*x-3,  unexpected decision
 : pol1_2    -- CONST -3 * ++  ;   x -> -3*x+4, also surprised!
 : pol1_3    --  ;                 x -> x-1
 : pol1_4    DUP + --  ;           x -> 2*x-1
 : pol1_5    CONST 10 * -- -- -- ; x -> 10*x-3, Cost without "3"!
 : pol1_6    CONST 11 * CONST 7 + ;x -> 11*x + 7
 : pol2_1    DUP OVER + *  ;       x -> 2*x*x
 : pol2_2    DUP CONST -3 * ++ * ; x -> -3*x*x + x
 : pol2_3    -- DUP *  ;           x -> x*x -2*x + 1
\end{verbatim}

Polynomials with division:

\begin{verbatim}
 : pol2d_1    DUP ++ * CONST 3 -- / --  ;  Quadratic: x->(x^2 + x -2)/2
\end{verbatim}

\bigskip

{\bf Working time}. Having $33$ basic words, we get $\approx 10^9$
programs of length $6$. If $\approx 10$ million programs per second
are processed (one core, frequency $\approx 3$ GHz), then a full
search will take about a minute and a half.

In an hour, you can go through all the programs of length $7$.

The search of all programs of length $8$ requires more than a day.

For an 8-core processor, it is realistic to check programs of length
$9$.

But this is not genetics and not biology. It is chemistry, and
inorganic.

One can increase the length of the programs by reducing the list of
basic words. However, at this stage it is not easy to do it
automatically.

\section{Probabilistic and Markov approaches}

\subsection{Partial programs}

In the simplest cases, as a result of a full search, we find a
program that performs all the specified tests. In more complicated
case to find solution only by a full search is impossible. However,
we obtain a list of the programs that perform part of the tests, for
example, the "factorial"\, of:
\begin{verbatim}
  3 :  DUP CONST 3 * +  ;
  4 :  DUP * CONST 5 -- / --  ;
  4 :  NEG DUP * -- CONST 5 /  ;
  3 :  DUP * DUP + ;
   ...
\end{verbatim}

The number at the beginning of the line indicates the number of
tests the program performs. Such programs will be called "partial
programs", their list is the main result of a complete search of
programs of small length.

All partial programs need not be memorized: a program that performs
a single test is hardly of interest. In our implementation, we set
the maximum number of partial programs ($N_p=400$ by default) and
remember only $N_p$ of the best, that is, those that perform the
largest number of tests, are stored.

In addition, a histogram is calculated showing how many times each
test was performed:
\begin{verbatim}
  0:  245594772
  1:     508224
  2:     294443
  3:      87869
  4:        412
  5:          0
  6:          0
  ...
\end{verbatim}

If after a full search in the list of partial programs some basic
words are not found at all or very rare, you can restrict the list
of basic words and repeat the full search with a limited set of
valid words to programs of longer length.

\subsection{Frequency table}

Having a list of partial programs, we will construct a list of
frequencies for the appearance of built-in words and the frequencies
of the appearance of word pairs.

\begin{remark}
In our algorithm, words and pairs that have never met, still receive
a small probability of $ (1/1024) $.
\end{remark}

Based on the frequency table, two methods of generating programs are
implemented:
\begin{itemize}
\item Probabilistic, when the probability of occurrence of the next
word in programs is taken from the frequency table;
\item Markov, when the probability of the next word depends on
the previous one.
\end{itemize}

The Markov approach should be more efficient, but usually the
frequency table is not large enough to more or less accurately
determine the frequencies of all pairs. Therefore, we propose a
combination of probabilistic and Markov approaches.

The following is an obvious statement.

\begin{proposition}
Suppose you have a program $P$, consisting of the commands $P=c_1,
\dots, c_k$, the probability of the appearance of $c_i$ a is $p_i$.
Denote by $\varepsilon$ the number $\varepsilon=p_1\dots p_k$, that
is, the probability of the program $P$ and let
$N=round(1/\varepsilon)$. Then the probability of the appearance of
the program $P$ for $N$ tests is $\approx 63\%$, for $3N$ tests is
$\approx 95\%$, for $5N$ tests is $\approx 99.3\%$.

\end{proposition}

\subsection{Base step}

Base step with parameters $(L_0, L_1, T)$  in a fixed dictionary
will be called the following algorithm.

\begin{enumerate}
\item Performed a full search of programs up to length $\le L_0$.
Build a list of partial programs.
\item Using the partial programs, we build a list of word
frequencies and a list frequencies of word pairs.
\item Generate within a specified interval of time $T$ seconds of a random
program of length $L_0+1 \dots L_1$, consistently, first a simple,
then a Markov method.
\item We correct the frequency tables and repeat step 3, this
cycle is performed $8$ times.
\end{enumerate}

The typical values of the parameters for working on one processor
core is $(7,14,400)$, that is, first perform a full search of
programs of length $\le 7$, then $8$ times for $50$ seconds is
performed alternately probabilistic/Markov search of programs of
length from $8$ to $14$ with correction of the frequency table after
each cycle.

\subsection{Example: discriminant}

The method of probabilistic search can find the implementation of
many algorithms. Consider, for example, the discriminant: $(a,b,c)
\rightarrow b^2-4ac$. Full search of the result was not given, but
allowed to reduce the number of basic words (the size of the
"genome")
 {\small
\begin{verbatim}
 CONST DUP DROP SWAP OVER ROT -ROT NEG + - * ++ --
\end{verbatim}
}

On the basis of the constructed frequency table, a random search for
programs of length 9 was started. Quite quickly the solution was
found: {\small
\begin{verbatim}
: discr OVER ROT * -ROT * CONST 4 * -  ;
\end{verbatim}
}

\subsection{Arbitrary (random) functions}

Let $ N $ be a natural number, denote by $X$ the set of $N$ elements
$X = \{0, \dots, N-1 \}$. We consider all possible functions $f:
X\rightarrow X$. Each such function is fully described by a set of
$N$ tests:
\[
\begin{array}{ll}
 0 & \rightarrow f(0) \\
 1 & \rightarrow f(1) \\
  & \ldots \\
 N-1 & \rightarrow f(N-1) \\
\end{array}
\]

The number of different functions is $N^N$.
\[
\begin{array}{ll}
 N & N^N \\
 3 & 27 \\
 4 & 256 \\
 5 & 3125 \\
 6 & 46 656 \\
\end{array}
\]

At $N=3$, all 27 functions have an implementation of the length $\le
5$, which is found by a full search, for example( the numbers at the
beginning of the string mean the values $f (0), f (1), f(2)$,
respectively):

\begin{verbatim}
 0 2 1 :  DUP 0> ++ OR --  ;
 1 0 2 :  DUP 0= ++ OR --  ;
 2 0 1 :  -- DUP 0< ++ AND  ;
 2 2 0 :  CONST -3 OR NEG --  ;
\end{verbatim}

At $N=4$ of $256$ functions, about 200 have an implementation of the
length $\le 6$. For all the checked functions of the remaining,
using probabilistic methods, an implementation of lengths from $7$
to $12$ was found, for example:

\begin{verbatim}
 0 3 1 0: DUP -- -- 0= 0= SWAP -- 0= ++ XOR ;  (len=10)
 0 3 1 1: DUP -- 0= DUP ++ ROT 0= - + ;        (len= 9)
 3 2 0 3: -- DUP -- 0= 0= ++ SWAP 0= 0= XOR ;  (len=10)
 3 2 3 1: DUP DUP ++ OR XOR ++ ++ DUP -- -- / ;(len=11)
 3 3 2 0: ++ DUP 0= 0= ++ /% ++ ++ XOR ;       (len= 9)
 3 3 3 0: DUP -- AND DUP ++ DUP OR -- 0= ++ ++ XOR ;(len=12)
\end{verbatim}

However, here there is no guarantee that the length is minimal. For
example, the last program can be replaced with
\begin{verbatim}
 :  --  --  --  0< CONST 3 * ;     (len=7)
\end{verbatim}

\subsection{Right shift}

Consider the function $(n,x) \rightarrow 2^n\,\cdot x$. A test file
containing $13$ test items was taken. Full search was unsuccessful,
but allowed to reduce the list of basic words ("genome") to: {\small
\begin{verbatim}
IF CONST DUP DROP SWAP OVER ROT -ROT NEG  +  -  --  ++  *
\end{verbatim}
}

After that, probabilistic search received as many as 5(!) programs
that perform $9$ tests (of $13$). They were very similar:
 {\small
\begin{verbatim}
  9:  DUP + SWAP -- DUP -ROT IF -7 +  ;
  9:  DUP ROT -- -ROT + OVER IF -7 +  ;
  9:  SWAP -- OVER ROT + OVER IF -7 +  ;
  9:  SWAP -- SWAP DUP + OVER IF -7 +  ;
  9:  DUP + SWAP -- SWAP OVER IF -7 +  ;
\end{verbatim}
}

And what tests are not performed? It turns out that the simplest,
when $n=0$: {\small
\begin{verbatim}
 <in> 0   3 </in><out>     3 </out>
 <in> 0   5 </in><out>     5 </out>
 <in> 0   5 </in><out>     5 </out>
 <in> 0   7 </in><out>     7 </out>
\end{verbatim}
}

Here you have a ready gene!

\section{Genetics}

In the biological interpretation of our results, the probabilistic
and Markov approach can be considered as "organic chemistry". Let's
take one more step and move on to "genetics".

As already mentioned, the list of words in the dictionary plays the
role of "genome". Each gene added can be either "good", "useful", or
"unsuccessful", "useless". The quality of the function-gene itself
can not be determined, only as part of the genome, aimed at solving
a specific problem. Therefore, we need to determine the quality of
the genome (dictionary)  in relation to this set of tests. If the
new gene improves the quality, we will consider it a good one.

Since our goal is to find a program that passes all the tests, the
quality it would naturally to consider the largest number of tests
that one of the programs passed. When this value is equal to the
total number of tests, the answer is found!

However, in practice this definition of quality is insufficient.
Often even a obviously successful gene does not increase the maximum
number of tests passed: it was $3$ and there remained $3$.

The "quality" of the genome will be the average number of tests
performed for one million randomly generated programs.

The new gene will be called "successful" if the maximum number of
tests performed has grown or remained the same, but the quality of
the genome has increased by at least $1$ percent.

As the "candidates for genes" we will take the most frequent chains
of bytes of length $2$ or $3$ among partial programs.

\subsection{Selecting a list of admissible words}\label{admit}

The same idea can be used to select a list of admissible words.

\begin{enumerate}
\item First, we take the minimum possible list of basic words ((11 words) as admissible.
\item In turn, we add one system word to this list.
\item Perform the basic step, find the quality of the genome (dictionary).
\item If the gene is successful, leave it in the list, otherwise delete it.
\end{enumerate}

\subsection{General algorithm}
During the process we will have two lists of programs (genes):
"candidates for genes"\, and "unsuccessful genes". There should be
no duplicate elements in them.

\begin{enumerate}
\item Find a list of valid words. Upon completion of the search,
we get, besides this list, also a list of partial programs.
\item Clear the list "candidates for the genes" and "bad genes".
\item In the lists of "candidates for genes" add the most frequent
chains of length $2$, $3$, and only those that are not in the list
of "unsuccessful genes".
\item If there are no candidates for genes, we finish the job.
\item Add one of the candidates to the dictionary.
\item Perform a basic step with this genome, find its quality.
\item If the gene was unsuccessful, remove it from the dictionary
and add it to the list of "unsuccessful genes".
\item Goto item (3).
\end{enumerate}

\section{Results}

Let's describe the results obtained with the help of the developed
program.

\subsection{Program description}

The program {\tt gene.exe} along with the detailed description is
available on the site \cite{gene_exe}. Its main features:
\begin{itemize}
\item setting a list of basic and additional words;
\item full search of programs up to length $L$ inclusive;
\item full search of programs within no more than $T$ seconds;
\item search probabilistic for $T$ seconds;
\item Markov search for $T$ seconds;
\item full search to the length $L$ and the probabilistic/Markov search
to the length $L1$ within $T$ seconds;
\item search for valid words.
\end{itemize}

\subsection{Digits}

The lowest digits were successfully found by direct search:

\begin{verbatim}
 : digit0     CONST 10 %  ;
 : digit1     CONST 10 / CONST 10 %  ;
 : digit2     CONST 10 DUP * / CONST 10 %  ;
\end{verbatim}

To find the remaining algorithms, a list of basic words was found in
which the constant $10$ was added:
\begin{verbatim}
: C10 CONST 10 ;
IF DUP DROP SWAP OVER ROT -ROT / % -- 0= C10
\end{verbatim}

The 3rd digit was found by probabilistic search. Here is the result
(of length 8):
\begin{verbatim}
: digit3 C10 / C10 / C10 / C10 %  ;
\end{verbatim}

Similarly, the highest digit was found (program of length 9):
\begin{verbatim}
: digitH DUP C10 / DUP -ROT IF -5 SWAP DROP  ;
\end{verbatim}

But the k-th figure in this way could not be found. To find the
program, a genetic approach has already been needed. During the
search, the genes were checked: {\small
\begin{verbatim}
 : F_036 DROP C10  ;
 : F_036 SWAP DROP  ;
 : F_036 % SWAP  ;
 : F_037 C10 %  ;
 : F_038 DROP C10 %  ;
 : F_039 SWAP DROP C10  ;
 : F_039 % SWAP DROP  ;
 : F_039 C10 % SWAP  ;
 : F_039 C10 F_036  ;
 : F_039 F_036 DROP  ;
 : F_039 C10 F_036 DROP  ;
 : F_039 C10 F_036 0=  ;
\end{verbatim}
}

"Useful"\, of them were recognized: {\small
\begin{verbatim}
 : F_036 % SWAP  ;
 : F_037 C10 %  ;
 : F_038 DROP C10 %  ;
 : F_039 C10 F_036 0=  ;
\end{verbatim}
}

Finally, the answer of length 12 was found: {\small
\begin{verbatim}
: digitK DUP IF 4 SWAP C10 / OVER -- ROT IF -7 F_038  ;
\end{verbatim}
}

As you can see, from the "useful"\, genes it was used only one {\tt
f\_038}. If "open" the gene, we get the program of length 14:
{\small
\begin{verbatim}
: digitK DUP IF 4 SWAP C10 / OVER -- ROT IF -7 DROP C10 % ;
\end{verbatim}
}

\subsection{Factorial}

For the function "factorial", $ n \rightarrow n!$, take a test file
consisting of $13$ tests for $n \le 13$. We start the search for a
set of admissible words and get it:
\begin{verbatim}
GOTO IF CONST DUP DROP SWAP OVER ROT -ROT + - * = --
\end{verbatim}

After that we try to build the required program.

Unfortunately, it is not possible to do this directly. For better
understanding, let's throw out the test line $1!=1$. On the
remaining set, start the search for the program by genetic method.

Because the process is probabilistic, different launches will
produce different results.

At the first start, the word appeared as a gene
\begin{verbatim}
: F_035 DUP DUP ;
\end{verbatim}
and already at the next step was found the program that performs all
the tests:
\begin{verbatim}
: FACTORIAL_1 F_035 IF 5 DUP ROT * SWAP -- DUP IF -7 IF -12  ;
\end{verbatim}

At the second start, the word appeared as a gene $F\_035$, but the
finished program could not be found at once. Two additional genes
were first constructed:
\begin{verbatim}
: F_036 DUP --  ; : F_037 -- DUP  ;
\end{verbatim}
and only after this the result is obtained:
\begin{verbatim}
: FACTORIAL_1 F_036 DUP -ROT * SWAP F_037 IF -6 +  ;
\end{verbatim}

If you substitute intermediate genes, insert them in the final
program, we get:
\begin{verbatim}
: FACTORIAL_1 DUP DUP IF 5 DUP ROT * SWAP -- DUP IF -7 IF -13 ;
\end{verbatim}
in the first case (length 14)
\begin{verbatim}
: FACTORIAL_1 DUP -- DUP -ROT * SWAP -- DUP IF -7 +  ;
\end{verbatim}
in the second (length 11).

And the hand-made program:
\begin{verbatim}
: FACTORIAL_1 DUP -- SWAP OVER * SWAP -- DUP IF -7 DROP ; (len=11)
\end{verbatim}

Add a deleted item to the tests:
\begin{verbatim}
<in> 1</in><out>      1 </out>
\end{verbatim}
and an additional gene (FACTORIAL\_0). After this, the desired
program is already found:
\begin{verbatim}
: FACTORIAL_0 DUP -- 0= IF 2  FACTORIAL_1  ;
\end{verbatim}

If we substitute genes, we get a result of length $16$:

\begin{verbatim}
 : FACTORIAL_0 DUP -- 0= IF 12 DUP -- SWAP OVER * SWAP -- DUP IF -7 DROP  ;
\end{verbatim}

\section{Genetic-2}

The results obtained above can be considered as bottom-up genetics,
that is, the construction of small genes, from which the final
program is built. You can consider another approach, when the genes
taken as a whole program that performs some task. Unfortunately,
this process has not been fully automated yet. Therefore, we will
look at it in examples, when genes are constructed in a
"semi-manual" mode.

\subsection{GCD, LCM}

Consider a test file GCD.tst of 180 elements:

\begin{verbatim}
#T GCD  x,y-> GCD(x,y)
 <in>     4      4 </in><out>     4 </out>
 <in>     5      5 </in><out>     5 </out>
 <in>     0      9 </in><out>     9 </out>
 <in>     7      0 </in><out>     9 </out>
 <in>     2     10 </in><out>     2 </out>
 <in>     4     10 </in><out>     2 </out>
 <in>     4     12 </in><out>     4 </out>
 <in>    12      5 </in><out>     1 </out>
 <in>    12      3 </in><out>     3 </out>
        ...
 <in>-33964  90856 </in><out>     4 </out>
 <in> 38817 -90856 </in><out>     1 </out>
\end{verbatim}

Brute force can not find a solution. The probabilistic method give a
partial program
\begin{verbatim}
: GCD_0  DUP -ROT % DUP IF -5 -  ;
\end{verbatim}
which gives the correct answer for all positive $(x, y)$. Adding it
as a gene, we find a partial program
\begin{verbatim}
: GCD_1 OVER ROT IF 3 GOTO -5 GCD_0  ;
\end{verbatim}
which gives the correct answer for all non-negative $(x,y)$. Adding
it as a gene and a abs-program
\begin{verbatim}
: ABS DUP 0> IF 2 NEG ;
\end{verbatim}
we fine the correct result:
\begin{verbatim}
: GCD ABS SWAP ABS GCD_1  ;
\end{verbatim}

On the basis of the found genes can easily be able to find an
algorithm for LCM:
\begin{verbatim}
: LCM ABS OVER OVER GCD / * ABS ;
\end{verbatim}
In fact, this algorithm will give an error for the pair (0.0). We
will not pay attention to this.

\subsection{What we want to get (Fibonacci)}

On the example of Fibonacci numbers.

We will look for a program that calculates Fibonacci numbers. Now,
even with the help of a genetic approach, this does not succeed.
Let's try to manipulate the genes "manually" to understand how this
should look in a more general situation.

We introduce an additional gene from the side, "from space". (see p.
\ref{sZakl} in next section): {\small
\begin{verbatim}
: fib1 -- -ROT SWAP OVER + ROT ;  (x,y,k) -> (y, x+y, k-1)
\end{verbatim} }

With its help we find the program (length 12): {\small
\begin{verbatim}
: Fibonacci0 ONE ONE ROT -- -- fib1 DUP IF -3 DROP SWAP DROP ;
\end{verbatim} }

It only works at $n \ge 3$. The general program will be as follows:
{\small
\begin{verbatim}
: Fibonacci3 DUP CONST 2 > IF 4 0> GOTO 2  Fibonacci0 ;
\end{verbatim} }

After expanding get (length 27):  {\small
\begin{verbatim}
: Fibonacci DUP CONST 2 > IF 4 0> GOTO 19 CONST 1 DUP ROT
   -- -- -- -ROT SWAP OVER + ROT DUP IF -8 DROP SWAP DROP ;
\end{verbatim} }

Thus, we can implement very complex algorithms, although the
automatic implementation of some steps looks rather problematic.

\section{Conclusion and future work}\label{sZakl}

Unfortunately, the developed methods do not allow us to find the
implementation of many algorithms. Some areas for further research
can be proposed now. Under this proposals, we can hope to make
significant progress in our direction.

\begin{enumerate}
\item Preparation of program templates.
\item Ranking of test items.
\item Frequency of appearance of triples.
\item Initial filling of the frequency table.
\item Saturation check for probabilistic search of a given length.
\item Initial genes "from space".
\item Setting the algorithm for selecting candidates for genes.
\item Transition to multi-core processor.
\item Merge multiple test files into one and pre-process it.
\item Efficient generation of structural programs.
\item Genetic-2.
\end{enumerate}

More detailed.

{\bf Preparation of program templates}. During the search, most of
the algorithms are discarded due to incorrect syntax, invalid
signature, etc. It is suggested to compile a complete list of
templates of syntactically correct programs with this signature in
advance and then select only from this list.

{\bf Ranking of test items}. Now, when building a list of partial
programs, all the test items are equivalent. This is not entirely
correct. One test is performed by very many programs, and the other
by a few units. Therefore, for each item, we consider how many
generated programs it is performed and in evaluating the quality of
the following programs we give preference to those that perform more
complex (rare) test items.

{\bf Frequency of appearance of triples}. At present, probabilistic
generation of programs takes into account only the probabilities of
occurrence of a single word and the probability of the appearance of
pairs. Practice shows that some triples have a rather high frequency
of appearance, noticeably greater than many pairs. Therefore, for a
probabilistic approach, such triples should be taken into account
separately.

{\bf Initial filling of the frequency table}. Now the frequency
table is built entirely on the basis of a partial programs. List of
these programs can be relatively small, so the frequencies of
individual words and especially pairs are too rough. It makes sense
to construct an initial filling of the frequency table of built-in
words and their pairs on the basis of an analysis of a large number
of real examples.

{\bf Saturation check for probabilistic search of a given length}.
At random/Markov program search of a short length ($8,9,10$) after a
while the generated programs begin to repeat. If such a phenomenon
is found, it should be assumed that programs of a given length have
already been exhausted and this length is not further considered.

{\bf Initial genes "from space"}. When discussing the problem of the
appearance of life on Earth, one of the hypotheses is the
introduction of life from space. In our case, this means the
appearance of genes from an external source.

More precisely, to the original 33-th built-in words add a few
dozens or even hundreds of additional words. If you use the
algorithm of finding admissible words (paragraph \ref{admit}), the
initial dictionary will not grow too much. Further work with such a
dictionary can be much more effective.

In this case, most likely, it is worth combining the basic words in
some groups with a fixed frequency ratio within the group. Each
group of words should be included in the basic dictionary only in
its entirety and determined only by the general probability of the
occurrence of a word from the group. The ratio of the frequencies
within the group should remain unchanged.

{\bf Setting the algorithm for selecting candidates for genes}.

Now candidates for genes are several of the most frequent substrings
of length $2$ and $3$ in the list of partial programs. In fact, it
is necessary to select substrings of different length, up to $5-6$,
but taking into account the quality of each partial program.

{\bf Transition to multi-core processor}. This is an extensive
approach, but now (2018-07-01) there are $28$ and $32$-core
processors. If not even consider supercomputers, such an increase in
processing power can lead to a noticeable increase in results.

{\bf Merge multiple test files into one and pre-process it}.
Sometimes it turns out to be impossible to find a program for this
test, genetic algorithms do not provide a sufficiently powerful
genome. In this case, a more complex test can help, which includes
several separate tests, but is related to a common theme. Then genes
built with the implementation of one test will help to advance and
implement another.

{\bf Efficient generation of structural programs}. Now with the
random generation of programs up to the actual execution of the
program on tests reaches about one program out of thousands, the
rest are discarded before that on various checks.  Based on this,
you can try to immediately generate only the correct program. This
will greatly increase the complexity of the generation algorithm,
but the potential gain in speed is worth it.

{\bf Genetic-2}

Now we're adding to the genome some of the frequently occurring
sub-strings in partial programs, this approach can be considered as
genetics "bottom-up", we start with simple programs and build more
complex ones from them. You can try to consider another approach,
"up-bottom".

\begin{definition}
Let the partial program $P$ be performed on a certain subset $S'$ of
all test items $S$. Let us denote a new test in $S_p$, in which the
input values coincide with the input values of the elements $S$, and
the return value is only one and is equal to $0$ or $1$, and $1$ on
a subset of $S'$ and $0$ on its addition.
\end{definition}

\begin{definition}
A partial program $P$ is said to be good if:
\begin{itemize}
\item It performs a significant part of all tests;
\item It is possible to find the program $Q_P$ realizing the test $S_P$.
\end{itemize}
\end{definition}

Let $P$ be some good partial program for the $S$ test. We denote by
$S''$ a test consisting of items on which $P$ NOT holds. Suppose
that we managed to find the program $R$ realizing the test $S''$.
Then the program that implements the original $S$ test looks like
this:

\begin{verbatim}
 : P2 Q IF 4 P GOTO 2 R ;
\end{verbatim}

\subsection{Conclusion}

The proposed method of genetic construction of programs allows us to
move quite far in this direction. Namely, we can implement programs
for algorithms sorting of $2$ and $3$ elements, finding the maximum
and minimum, many general and individual polynomials of one, two and
three variables, binary shifts, decimal and binary digits, GCM, LCM,
find a minimal prime divisor of a number and many others. However,
not all algorithms can be implemented. Further development of the
proposed methods should lead to a significant increase in the
capabilities.

In addition to the theoretical interest, you can already see its
practical application. For example, taking the assembly language of
a certain processor instead of the Forth language, you can optimize
individual sections of the program.



\end{document}